\documentclass[final]{cvpr}

\usepackage{times}
\usepackage{epsfig}
\usepackage{graphicx}
\usepackage{amsmath}
\usepackage{amssymb}

\usepackage{algorithm}
\usepackage{algorithmic}
\usepackage{multirow}
\usepackage{latexsym}
\usepackage{subfigure}
\usepackage[pagebackref=true,breaklinks=true,colorlinks,bookmarks=false]{hyperref}

\def\cvprPaperID{5343} 
\def\confYear{CVPR 2021}
\begin{document}

\title{Prototype Completion with Primitive Knowledge for Few-Shot Learning}

\author{Baoquan Zhang, Xutao Li\thanks{Corresponding author}, Yunming Ye\footnotemark[1], Zhichao Huang, Lisai Zhang\\	
Harbin Institute of Technology, Shenzhen\\
{\tt\small zhangbaoquan@stu.hit.edu.cn, \{lixutao, yeyunming\}@hit.edu.cn,}\\
{\tt\small iceshzc@stu.hit.edu.cn, LisaiZhang@foxmail.com}
}
\maketitle

\thispagestyle{empty} 
\begin{abstract}
   Few-shot learning is a challenging task, which aims to learn a classifier for novel classes with few examples. Pre-training based meta-learning methods effectively tackle the problem by pre-training a feature extractor and then fine-tuning it through the nearest centroid based meta-learning. However, results show that the fine-tuning step makes very marginal improvements. In this paper, 1) we figure out the key reason, \emph{i.e.}, in the pre-trained feature space, the base classes already form compact clusters while novel classes spread as groups with large variances, which implies that fine-tuning the feature extractor is less meaningful; 2) instead of fine-tuning the feature extractor, we focus on estimating more representative prototypes during meta-learning. Consequently, we propose a novel prototype completion based meta-learning framework. This framework first introduces primitive knowledge (\emph{i.e.}, class-level part or attribute annotations) and extracts representative attribute features as priors. Then, we design a prototype completion network to learn to complete prototypes with these priors. To avoid the prototype completion error caused by primitive knowledge noises or class differences, we further develop a Gaussian based prototype fusion strategy that combines the mean-based and completed prototypes by exploiting the unlabeled samples. Extensive experiments show that our method: (\romannumeral1) can obtain more accurate prototypes; (\romannumeral2) outperforms state-of-the-art techniques by $2\% \sim 9\%$ in terms of classification accuracy. Our code is available online \footnote {\url{https://github.com/zhangbq-research/Prototype\_Completion\_for\_FSL}}. 
\end{abstract}

\section{Introduction}
\label{sec1}
Humans can adapt to a novel task from only a few observations, because our brains have the excellent capability of learning to learn. In contrast, modern artificial intelligence (AI) systems generally require a large amount of annotated samples to make the adaptations. However, preparing sufficient annotated samples is often laborious, expensive, or even unrealistic in some applications, for example, cold-start recommendation \cite{VartakTMBL17} and drug discovery \cite{Altae-TranRPP16}. To equip the AI systems with such human-like ability, few-shot learning (FSL) becomes an important and widely studied problem. Different from conventional machine learning, FSL aims to learn a classifier from a set of base classes with abundant labeled samples, then adapt to a set of novel classes with few labeled data
\cite{WangYKN20}. 

\begin{figure}[t]
	\centering
	\subfigure[Base Classes ($\mathrm{\sigma^2=0.086}$)]{ 
		\label{fig1a} 
		\includegraphics[width=0.46\columnwidth]{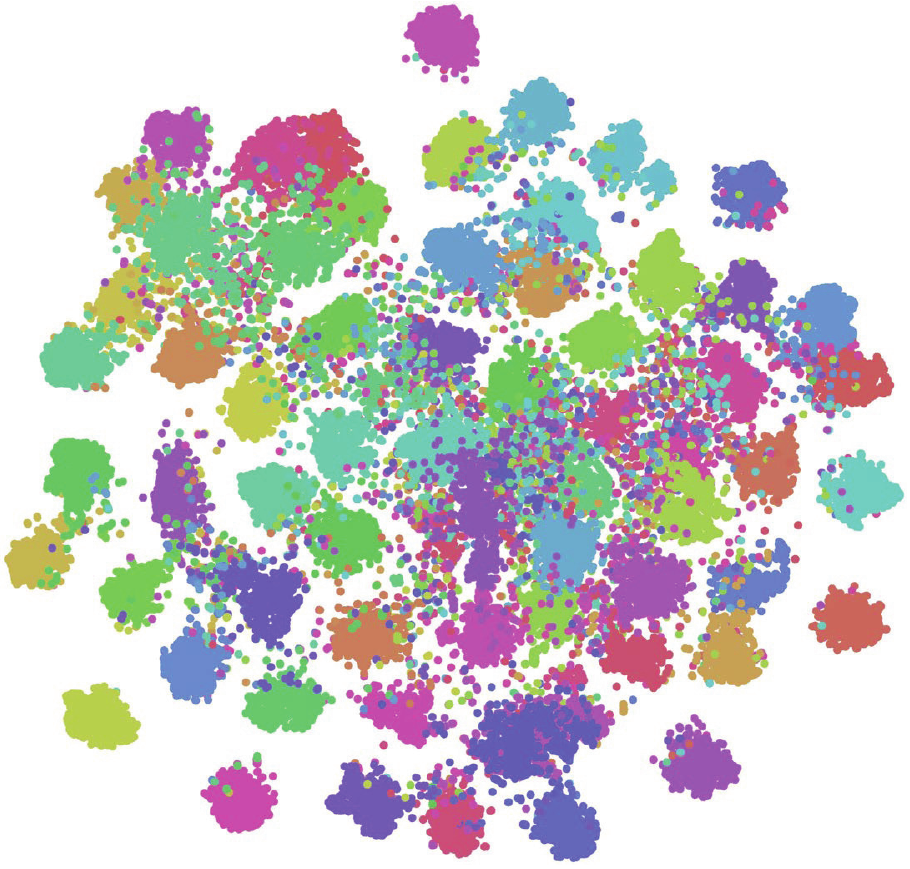}}
	\subfigure[Novel Classes ($\mathrm{\sigma^2=0.099}$)]{ 
		\label{fig1b} 
		\includegraphics[width=0.46\columnwidth]{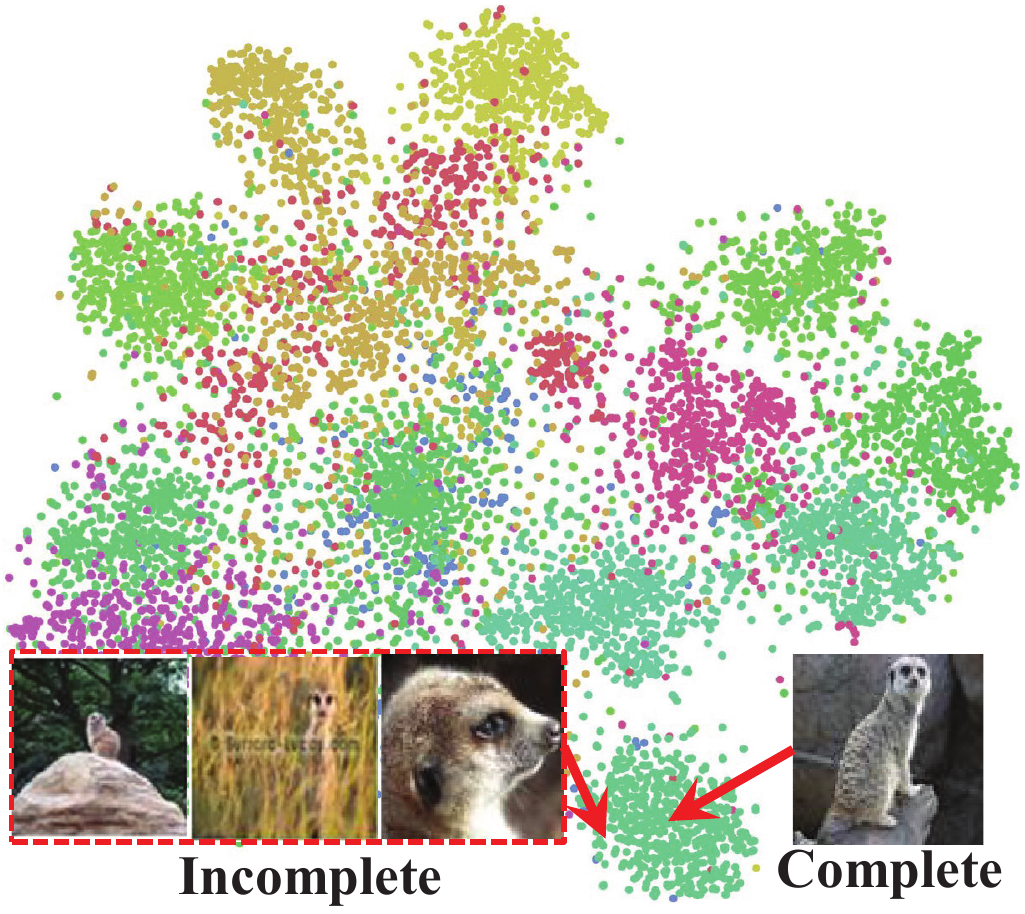}}
	\caption{The distribution of base and novel class samples in the pre-trained feature space. ``$\mathrm{\sigma}^2$'' denotes the averaged variance. }
	\label{fig1}
\end{figure}

Existing studies on FSL roughly fall into four categories, namely the metric-based methods \cite{ChenZWC20}, optimization-based methods \cite{finn2017model}, graph-based methods \cite{satorras2018few}, and semantics-based methods \cite{xing2019adaptive}. Though their methodologies are totally different, almost all methods address the FSL problem by a two-phase meta-learning framework, \emph{i.e.}, meta-training and meta-test phases. Recently, Chen \emph{et al. }\cite{chen2020new} find that introducing an extra pre-training phase can significantly boost the performance. In this method, a feature extractor first is pre-trained by learning a classifier on the entire base classes. Then, the metric-based meta-learning is adopted to fine-tune it. In the meta-test phase, the mean-based prototypes are constructed to classify novel classes via a nearest neighbor classifier with cosine distance. 

Though the pre-training based meta-learning method achieves promising improvements, Chen \emph{et al.} find that the fine-tuning step indeed makes very marginal contributions \cite{chen2020new}. However, the reason is not revealed in \cite{chen2020new}. To figure out the reason, we visualize the distribution of base and novel class samples of the miniImagenet in the pre-trained feature space in Figure~\ref{fig1}. We find that the base class samples form compact clusters while the novel class samples spread as groups with large variances. It means that 1) fine-tuning the feature extractor to gather the base class samples into more compact clusters is less meaningful, because this enlarges the probability to overfit the base tasks; 2) the given few labeled samples may be far away from its ground-truth centers in the case of large variances for novel classes, which poses a great challenge for estimating representative prototypes. Hence, in this paper, instead of fine-tuning the feature extractor, we focus on {\bf how to estimate representative prototypes from the few labeled samples}, especially when these samples are far away from its ground-truth centers. 

Recently, Xue \emph{et al.} \cite{XueW20} also attempt to address a similar problem by learning a mapping function from noisy samples to their ground-truth centers. 
However, learning to recover representative prototypes from noisy samples without any priors is very difficult. Moreover, the method does not leverage the pre-training strategy. Thus, the performance improvement of the method is limited. In this paper, we find that the samples deviated from its ground-truth centers are often incomplete, \emph{i.e.}, missing some representative attribute features. As shown in Figure~\ref{fig1b}, the meerkat sample nearby the class center contains all the representative features, \emph{e.g.}, the head, body, legs and tail, while the ones far away may miss some representative features. This means that the prototypes estimated by the samples deviated from its centers may be incomplete.

Based on this fact, we propose a novel prototype completion based meta-learning framework. Our framework works in a pre-training manner and introduces some primitive knowledge, e.g., whether a class object should have ears, legs or eyes, as priors to achieve the prototype completion. Specifically, we first extract the visual features for each part/attribute, by aggregating the pre-trained feature representations of all the base class samples that have the corresponding attribute in our primitive knowledge. Second, we mimic the setting of few-shot classification task and construct a set of prototype completion tasks. A \underline{Proto}type \underline{Com}pletion \underline{Net}work (ProtoComNet) is then designed to learn to complete representative prototypes with the primitive knowledge and visual attribute features. To avoid the prototype completion error caused by primitive knowledge noises or base-novel class differences, we further design a Gaussian-based prototype fusion strategy, which effectively combines the mean-based and completed prototypes by exploiting the unlabeled data. Finally, the few-shot classification is achieved via a nearest neighbor classifier. Our main contributions of this paper can be summarized as follows: 

\begin{itemize}
	\item We reveal the reason why the feature extractor fine-tuning step contributes marginally to the pre-training based meta-learning methods, and point out that representative prototype estimation is a more critical issue.
	
	\item We propose a novel prototype completion based meta-learning framework, which can effectively learn to recover representative prototypes by leveraging primitive knowledge and unlabeled data. 
	
	\item We have conducted comprehensive experiments on three real-world data sets. The experimental results demonstrate that the proposed method outperforms the state-of-the-art techniques by $2\% \sim 9\%$ in terms of classification accuracy.
\end{itemize}

\section{Related Work}
\subsection{Few-Shot Learning}
Meta-learning is an effective manner to solve the FSL problem. Existing approaches are mainly grouped into four categories. {\bf 1) Metric-based approaches.} The type of methods aim to learn a good metric space, where novel class samples can be nicely categorized via a nearest neighbor classifier with Euclidean \cite{snell2017prototypical} or cosine distance \cite{ChenLKWH19}. For example, Zhang \emph{et al.} \cite{ZhangZNXY19} attempted to learn the metric space by distribution based classification rules instead of point estimation. {\bf 2) Optimization-based approaches.} The methods follow the idea of modeling an optimization process over few labeled samples under the meta-learning framework, aiming to adapt to novel tasks by a few optimization steps, such as \cite{finn2017model, lee2019meta}.
{\bf 3) Graph-based approaches.} The methods learn how to construct a good graph structure and propagate the labels from base classes and then apply the meta-knowledge on novel classes \cite{liu2019learning, rodriguez2020embedding, satorras2018few}.
{\bf 4) Semantics-based approaches.} This line of methods employ the textual semantic knowledge to enhance the performance of meta-learning on FSL problems \cite{ChenFZJXS19, boostingfew}. For example, in \cite{zhimao2019few, babysteps2019, xing2019adaptive}, they explored the class correlations, respectively, from the perspectives of the class name, description, and knowledge graph as textual semantic knowledge, aiming to enhance the FSL classifier by the convex combination of visual and semantic modalities. Different from these works, we introduce fine-grained visual attributes to enable a meta-learner to learn to complete prototypes, instead of to combine two modalities.

\begin{figure*}[t]
	\centering
	\includegraphics[width=1.0\textwidth]{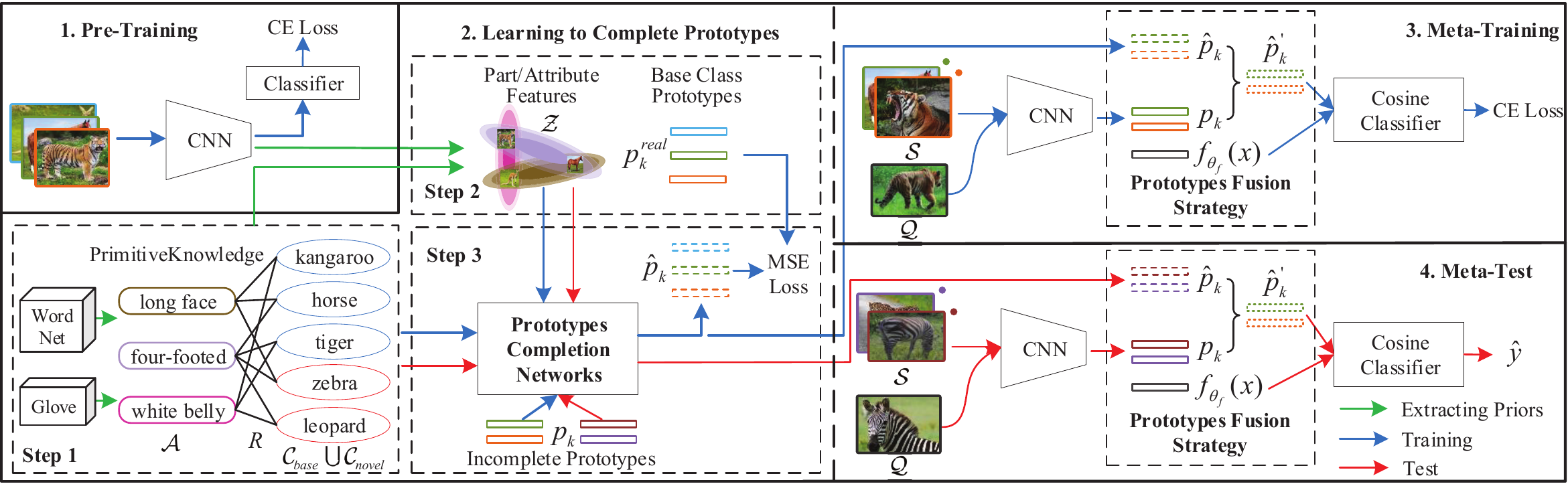} 
	\caption{The prototype completion based meta-learning framework.}
	\label{fig2}
\end{figure*}

Recently, some studies turn to pre-training techniques for the FSL problem and achieve promising performance. Chen \emph{et al.} \cite{ChenLKWH19} first proposed and investigated the pre-training techniques in FSL, by considering linear-based and cosine distance-based classifiers, respectively. Liu \emph{et al.} \cite{YaohuiWang_pr} developed a label propagation and feature shifting strategy to diminish the intra-class and cross-class bias of prototypes in the pre-trained feature space. In \cite{chen2020new}, a novel metric-based meta-learning method was developed by incorporating a pre-training phrase. These methods, albeit delivering promising performance, do not fully explore the power of pre-training, as results show that the major improvements are made by the pre-training, while the meta-learning phase contributes very marginally. According to our analysis, this is because novel classes group loosely in the pre-trained feature space. In such case, estimating a more accurate prototype is more important than fine-tuning the projection spaces. Hence, in this paper, we propose a prototype completion framework to address the issue.

\subsection{Zero-Shot Learning}
Zero-shot learning (ZSL) is also closely related to FSL, which aims to address the novel class categorizations without any labeled samples. The key idea is to learn a mapping function between the semantic and visual space on the base classes, then apply the mapping to categorize novel classes. The semantic spaces in ZSL are typically attribute-based \cite{WanCLYZY019}, text description-based \cite{ReedALS16}, and word vector-based \cite{FromeCSBDRM13}. For example, in \cite{WanCLYZY019}, the semantic attributes were employed and a structure constraint on visual centers was incorporated for the mapping function learning. Our method differs from those models in two key points: (\romannumeral1) our method is for the FSL problem, where few labeled samples should be effectively utilized; (\romannumeral2) relying on semantic attributes, we propose a novel prototype completion based meta-learning framework, instead of directly learning the map function.

\subsection{Visual Attributes}
Visual attributes refer to the visual feature of object components \cite{multilabelobjectattribute}, which have been successfully utilized in various domains, such as action recognition \cite{ZhangTGL18}, zero-shot learning \cite{WanCLYZY019}, person Re-ID \cite{LinZZWHYY19}, and image caption \cite{ChenDLZH18}. Recently, several FSL techniques relying on visual attributes have been proposed. In \cite{tokmakov2019learning}, an attribute decoupling regularizer was developed based on visual attributes to obtain good representations for images. Hu \emph{et al.} \cite{Hu2019Weaklys} proposed a compositional feature aggregation module to explore both spatial and semantic visual attributes for FSL. Zou \emph{et al.} \cite{zou2020compositional} explored compositional few-shot recognition by learning a feature representation composed of important visual attributes. All the methods utilize visual attributes for better representations. Different from these studies, we leverage them to learn a prototype completion strategy. As a result, more accurate prototypes can be obtained for FSL. 

\section{Methodology}

\subsection{Problem Definition}
For $N$-way $K$-shot problems, we are given two set: a training set $\mathcal{S}=\{(x_i, y_i)\}_{i=0}^{N \times K}$ with a few of labeled samples  (called the support set) and a test set $\mathcal{Q}=\{(x_i, y_i)\}_{i=0}^{M}$ consisting of unlabeled samples (called the query set). Here $x_i$ denotes the image sampled from the set of novel classes $\mathcal{C}_{novel}$, $y_i \in \mathcal{C}_{novel}$ is the label of $x_i$, $N$ indicates the number of classes in $\mathcal{S}$, $K$ denotes the number of images of each class in $\mathcal{S}$, and $M$ denotes the number of images in $\mathcal{Q}$. Meanwhile, we also have an auxiliary data set with abundant labeled images $\mathcal{D}_{base}=\{(x_i, y_i)\}_{i=0}^{B}$, where $B$ is the number of images in $\mathcal{D}_{base}$, the image $x_i$ is sampled from the set of base classes $\mathcal{C}_{base}$, and the sets of class $\mathcal{C}_{base}$ and $\mathcal{C}_{novel}$ are disjoint, \emph{i.e.}\ $y_i \in \mathcal{C}_{base}$ and $\mathcal{C}_{base} \cap \mathcal{C}_{novel} = \emptyset$. Our goal is to learn a good classifier for the query set $\mathcal{Q}$ on the support set $\mathcal{S}$ and the auxiliary dataset $\mathcal{D}_{base}$.

\subsection{Overall Framework}
\label{section_3_2}
As shown in Figure~\ref{fig2}, the proposed framework consists of four phases, including pre-training, learning to complete prototypes, meta-training and meta-test. 

\noindent {\bf Pre-Training.} In the phase, we build and train a convolution neural network (CNN) classifier with the base classes samples. Then, the last softmax layer is removed and the classifier turns into a feature extractor $f_{\theta_f}()$ with parameters $\theta_f$. This offers a good embedding representation. 

\noindent {\bf Learning to Complete Prototypes.} We propose a Prototype Completion Network (ProtoComNet) as a meta-learner. It accounts for complementing the missing attributes for incomplete prototypes. The main details of the ProtoComNet will be elaborated in Section~\ref{section3_3}. Here we first give an
overview of its workflow depicted in Figure~\ref{fig2}, which includes three steps: 

{\bf Step 1}. We construct primitive knowledge for all the classes. The knowledge is what kinds of attribute feature the class should have, \emph{e.g.}, the kangaroo has long face and white belly, and zebra has long face and four feet. We note that such kinds of knowledge is very cheap to obtain, \emph{e.g.}, from WordNet. 
Let $\mathcal{A} = \{a_i\}_{i=0}^{F}$ denotes the set of class parts/attributes where $F$ is the number of attributes, and $R$ denotes the association matrix between the attributes and the classes, where $R_{ka_i}=1$ if the attribute $a_i$ is associated with the class $k$; otherwise $R_{ka_i}=0$. Meanwhile, the semantic embeddings of all classes and attributes are calculated by Glove \cite{pennington2014glove} in an average manner of word embeddings, denoted by $\mathcal{H} = \{h_k\}_{k=0}^{|\mathcal{C}_{base}|+|\mathcal{C}_{novel}|-1} \cup \{h_{a_i}\}_{i=0}^F$. 

{\bf Step 2}. Based on the pre-trained feature extractor $f_{\theta_f}()$ and primitive knowledge, we extract two types of information, namely base class prototypes and part/attribute features. Specifically, the base class prototypes $p_k^{real}$ can be calculated by averaging the extracted features of all samples in the base class $k$, that is,
\begin{equation} 
p_k^{real}=\frac {1}{|\mathcal{D}_{base}^k|} \sum_{(x, y) \in \mathcal{D}_{base}^k} f_{\theta_f}(x),
\label{eq1}
\end{equation}
where $\mathcal{D}_{base}^k$ denotes the set of samples from the base class $k$. As for the feature $z_{a_i}$ of part/attribute $a_i$, our intuition is that it can be transfered from base classes to novel classes. For example, even if human haven't seen ``zebra'', they can also imagine its visual features of ``long face'' once they learn “long face” from ``kangaroo'' and ``horse''. To obtain the part/attribute feature $z_{a_i}$, we denote all base class samples that have the corresponding part/attribute $a_i$ in the primitive knowledge as a set $D_{base}^{a_i}$. Then, we calculate its mean $\mu_{a_i}$ and diagonal covariance $diag(\sigma_{a_i}^2)$ as: 

\begin{equation} 
\mu_{a_i}=\frac {1}{|\mathcal{D}_{base}^{a_i}|} \sum_{(x, y) \in \mathcal{D}_{base}^{a_i}} f_{\theta_f}(x),
\label{eq2}
\end{equation}
\begin{equation} 
\sigma_{a_i}=\sqrt{\frac {1}{|\mathcal{D}_{base}^{a_i}|} \sum_{(x, y) \in \mathcal{D}_{base}^{a_i}} (f_{\theta_f}(x)\ -\ \mu_{a_i})^2}.
\label{eq3}
\end{equation}
Here, the mean $u_{a_i}$ and the diagonal covariance $diag(\sigma_{a_i}^2)$ characterize the part/attribute feature distribution of attribute $a_i$, \emph{i.e.}, $z_{a_i} \sim N(\mu_{a_i}, diag(\sigma_{a_i}^2))$, which will be used in Section~\ref{section3_3}.

{\bf Step 3}. Upon the results of the previous steps, we mimic the setting of $K$-shot tasks and construct a set of prototype completion tasks to train our meta-learner $f_{\theta_c}()$ (\emph{i.e.}, ProtoComNet) in an episodic manner \cite{vinyals2016matching}. 
Specifically, in each episode, we randomly select one class $k$ from base classes $C_{base}$ and $K$ images for the class $k$ from $\mathcal{D}_{base}$ as support set $S$. Then, we average the features of all samples in $S$ as the incomplete prototypes $p_k$. Here, we consider it as incomplete because some representative features may be missing. Even though in some cases this may not be true, regarding them as incomplete ones does no harms to our meta-learner. 
Finally, we take the incomplete prototypes $p_k$, the primitive knowledge (the class-attribute association matrix $R$ and word embeddings $\mathcal{H}$), and the parts/attribute feature $\mathcal{Z}=\{z_{a_i}\}_{i=0}^{F}$ as inputs, and treat the base class prototypes $p_k^{real}$ as targets, to train our meta-learner by using the Mean-Square Error (MSE) loss. That is, 
\begin{equation} 
	\min\limits_{\theta_c} \mathbb{E}_{(p_k,\ p_k^{real}) \in \mathbb{T}} \ MSE(f_{\theta_c}(p_k, R, \mathcal{H}, \mathcal{Z}),\ p_k^{real}),
\label{eq4}
\end{equation}
where $\theta_c$ denotes the parameters of our meta-learner and $\mathbb{T}$ denotes the set of prototype completion tasks. 

\noindent {\bf Meta-Training.} 
To jointly fine-tune the feature extractor $f_{\theta_f}()$ and the meta-learner $f_{\theta_c}()$, we construct a number of $N$-way $K$-shot tasks from $\mathcal{D}_{base}$ following the episodic training manner \cite{vinyals2016matching}. Specifically, in each episode, we sample $N$ classes from the base classes $\mathcal{C}_{base}$, $K$ images in each class 
as the support set $\mathcal{S}$, and $M$ images as the query set $\mathcal{Q}$. Then, $f_{\theta_f}()$ and $f_{\theta_c}()$ can be further fine-tuned by maximizing the likelihood estimation on query set $Q$. That is,
\begin{equation}
\begin{aligned}
\max\limits_{\theta}  \mathbb{E}_{(\mathcal{S},\mathcal{Q}) \in \mathbb{T}'} \sum_{(x,y) \in \mathcal{Q}} log(P(y|x, \mathcal{S}, R, \mathcal{H}, \mathcal{Z}, \theta)),
\end{aligned}
\label{eq5}
\end{equation}
where $\theta = \{\theta_f, \theta_c\}$ and $\mathbb{T}'$ denotes the set of $N$-way $K$-shot tasks. Specifically, for each episode, we first estimate its class prototype $p_k$ by averaging the features of the labeled samples. That is,
\begin{equation}
p_k = \frac{1}{|\mathcal{S}_k|} \sum_{x \in \mathcal{S}_k} f_{\theta_f}(x),
\label{eq6}
\end{equation}
where $\mathcal{S}_k$ is the support set extracted for the class $k$. 
Then, the ProtoComNet is applied to complete $p_k$, and we have: 
\begin{equation}
\hat{p}_k = f_{\theta_c}(p_k, R, \mathcal{H}, \mathcal{Z}).
\label{eq7}
\end{equation}
Moreover, to obtain more reliable prototypes, we further explore unlabeled samples and combine $p_k$ and $\hat{p}_k$ by introducing a Gaussian-based prototype fusion strategy (which will be introduced in Section~\ref{section3_4}). As a result, the fused prototype $\hat{p}'_k$ is obtained. Finally, the probability of each sample $x \in \mathcal{Q}$ to be class $k$ is estimated based on the proximity between its feature $f_{\theta_f}(x)$ and $\hat{p}'_k$. That is,
\begin{equation}
P(y=k|x, \mathcal{S}, R, \mathcal{H}, \mathcal{Z}, \theta) = \frac{e^{d(f_{\theta_f}(x),\ \hat{p}'_k)\ \cdot\ \gamma}}
{\sum_{c} e^{d(f_{\theta_f}(x),\ \hat{p}'_c)\ \cdot\ \gamma}},
\label{eq8}
\end{equation}
where $d()$ denotes the cosine similarity of two vectors and $\gamma$ is a learnable scale parameter.

\noindent {\bf Meta-Test.} Following Eqs. (\ref{eq6}) $\sim$ (\ref{eq8}), we directly perform few-shot classification for novel classes. 

\subsection{Prototypes Completion Network}
\label{section3_3}
\begin{figure}[t]
	\centering
	\includegraphics[width=0.92\columnwidth]{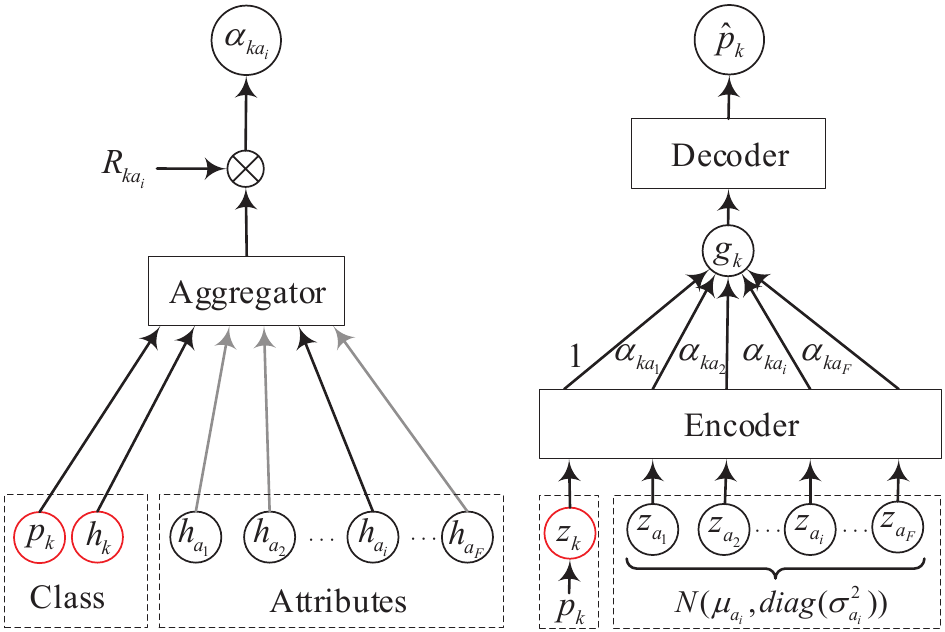} 
	\caption{Illustration of the encoder-aggregator-decoder networks. 
	}
	\vspace{-10pt}
	\label{fig3}
\end{figure}

In this subsection, we introduce how the ProtoComNet $f_{\theta_c}()$ is designed. Our notion is treating the primitive knowledge ($\cal R$ and $\cal H$), part/attribute features $\cal Z$ and the incomplete prototype $p_k$ as inputs and the completed prototype $\hat{p}_k$ as output, and then building an encoder-aggregator-decoder network, as shown in Figure~\ref{fig3}. The encoder aims to form a low-dimensional representation of estimated prototypes and part/attributes. Then, the aggregator accounts for evaluating the importance of different parts/attributes and combining them with a weighted sum. Finally, the decoder is in charge of the prediction of complete prototypes ${\hat p}_k$. Next, we detail the three components, respectively.

\noindent {\bf The Encoder.} In the training part, the encoding process involves a sampling of a class attribute feature $z_{a_i}$ from its feature distribution $N(\mu_{a_i}, diag(\sigma_{a_i}^2))$, followed by an encoder $g_{\theta_e}()$ that encodes the attribute feature $z_{a_i}$ and the estimated prototypes $p_k$ to a latent code $z'_{a_i}$ and $z'_k$, respectively. The overall encoding process is defined in Eq. (\ref{eq9}):
\begin{equation} 
\begin{aligned}
z_{a_i} \ \sim \ &N(\mu_{a_i}, \ diag(\sigma_{a_i}^2))
,\ z_{a_i}' = g_{\theta_e}(z_{a_i}), \\&
z_{k} \ = p_k
,\ z_{k}' = g_{\theta_e}(z_{k}),
\end{aligned}
\label{eq9}
\end{equation}
where $\theta_e$ denotes the parameters of the encoder. Note that we use the mean $\mu_{a_i}$ to replace $z_{a_i}$ in the meta-test phase.

\noindent {\bf The Aggregator.} Intuitively, different parts/attributes make varying contributions to distinct classes, for example, the ``nose'' is more representive for elephants than tigers to complete their prototypes. Hence, differentiating their contributions in the completion is important. To this end, we employ an attention-based aggregator $g_{\theta_a}()$. Here, we calculate the attention weights $\alpha_{ka_i}$ by using the semantic embeddings $h_{k}$ and $h_{a_i}$ of the class $k$ and the attribute $a_i$, and the incomplete prototype $p_k$. Then, we apply them to combine the latent codes $z_{k}'$ and $z_{a_i}'$, and obtain the aggregated result $g_k$ as follows: 
\begin{small}
\begin{equation} 
\begin{aligned}
\alpha_{ka_i}=R_{ka_i}g_{\theta_a}(p_{k} || h_{k} || h_{a_i})
,\ g_k = \sum_{a_i}\alpha_{ka_i}z'_{a_i}+z'_k,
\end{aligned}
\label{eq10}
\end{equation}
\end{small}where $\theta_a$ is the parameters of the aggregator and $||$ is a concatenation operation. 

\noindent {\bf The Decoder.} Finally, we use the aggregated result $g_k$ to decode the complete prototypes $\hat{p}_{k}$ for each class $k$ by the decoder module $g_{\theta_d}()$. That is, $\hat{p}_{k} = g_{\theta_d}(g_k)$, where $\theta_d$ denotes the parameters of the decoder. 

\subsection{Prototype Fusion Strategy}
\label{section3_4}
Till now, we have two prototype estimations, \emph{i.e.}, the mean-based prototype $p_k$ and the completed prototype ${\hat p}_k$. Next, we will discuss why and how to fuse these two estimations from the perspective of Bayesian estimation.

\begin{figure}[t]
	\centering
	\subfigure[Experiment on 5-way $K$-shot task]{ 
		\label{fig3_4_a} 
		\includegraphics[width=0.50\columnwidth]{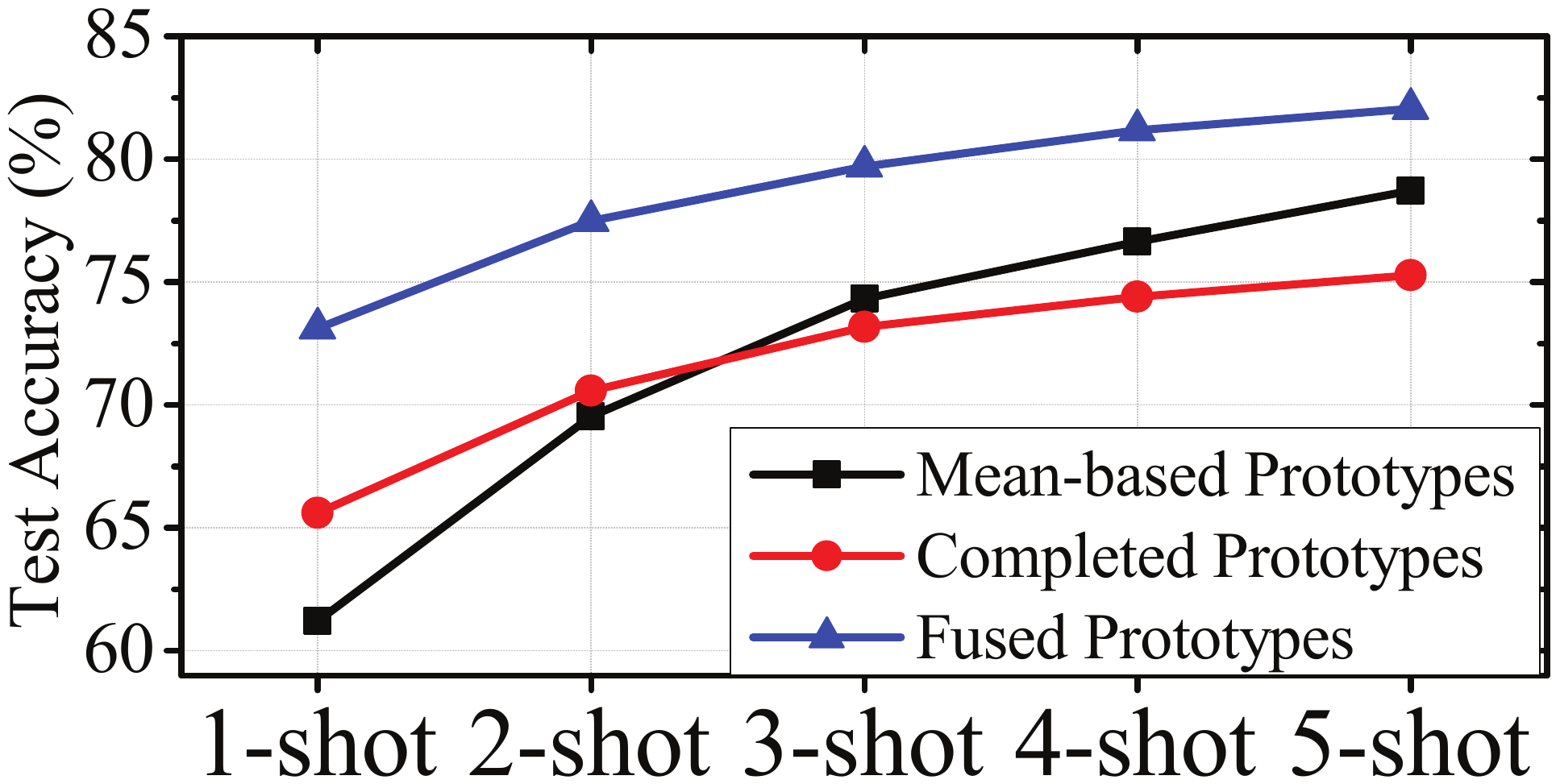}}
	\subfigure[Prototype fusion strategy]{ 
		\label{fig3_4_b} 
		\includegraphics[width=0.45\columnwidth]{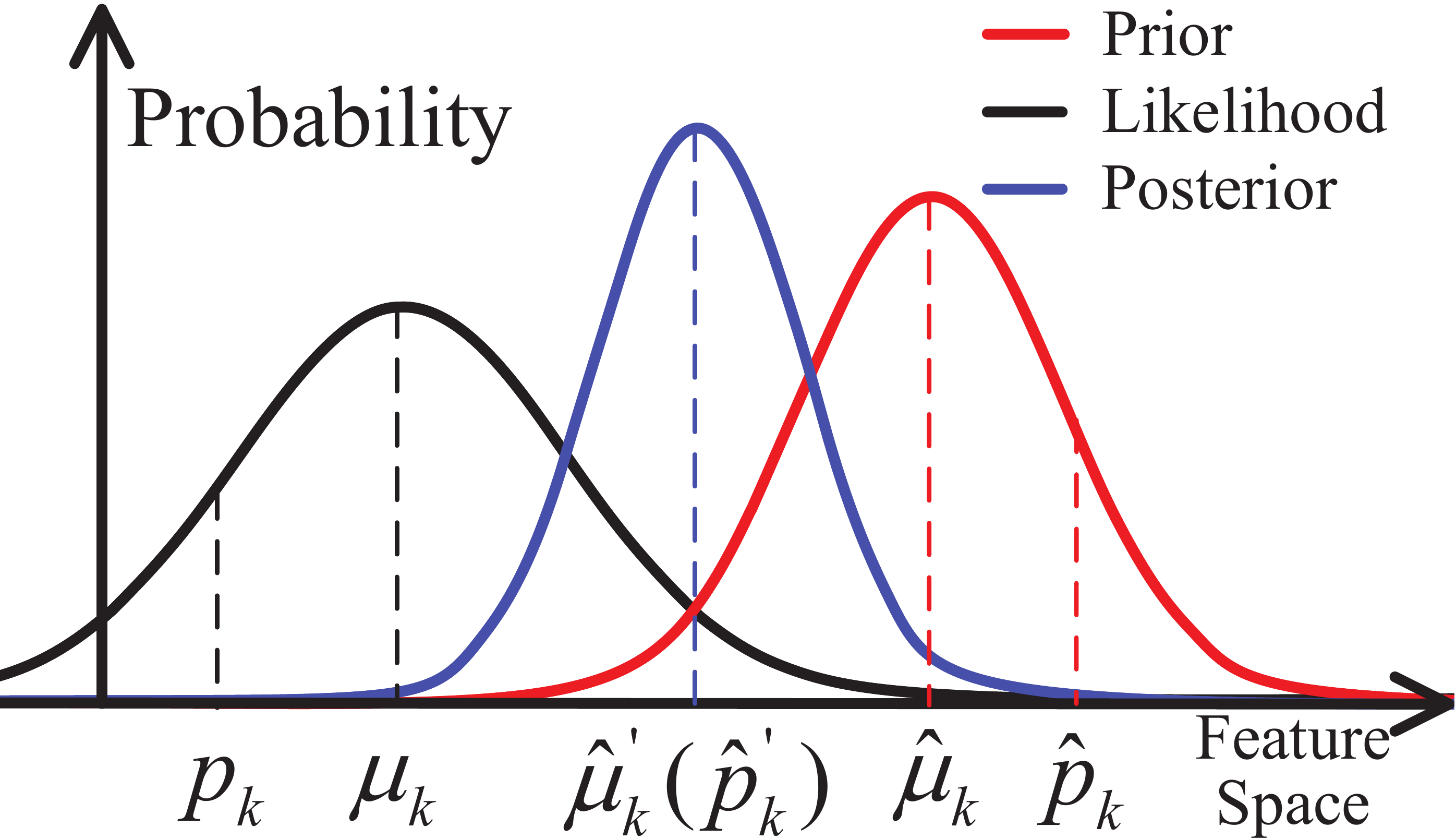}}
	\caption{Test accuracy of $p_k$ and ${\hat p}_k$ on 5-way $K$-shot tasks of miniImagenet (a) and Illustration of prototype fusion strategy (b). }
	\vspace{-10pt}
	\label{fig3_4}
\end{figure}

\noindent {\bf Why do we fuse prototypes?} Actually, both the estimates $p_k$ and ${\hat p}_k$ have their own biases. The former is mainly due to the scarcity or incompleteness of labeled samples in novel classes, which produces biased means; while the latter is brought by the primitive knowledge noises and the base-novel class differences. The fact implies that the two estimates can remedy each other. When the labeled samples are very scarce and incomplete, the completed prototype ${\hat p}_k$ is more reliable because the completion is learned from a great number of base class tasks. As more and more labeled samples become available, the mean-based prototype is more representative because the ProtoComNet may result in prototype completion error problem under the effects of primitive knowledge noises or class differences. Figure~\ref{fig3_4_a} shows an example to demonstrate this. We observe that the completed prototypes are more accurate on 1/2-shot tasks while the mean-based ones are better on 3/4/5-shot tasks. Thus, a prototype fusion strategy is desired to combine their advantages and form more representative prototypes.

\noindent {\bf How to fuse prototypes?} We apply the Bayesian estimation to fuse the two kinds of prototypes. Specifically, we assume that the estimated prototypes follow the Multivariate Gaussian Distribution (MGD), as the samples in the pre-trained space are continuous and clustered together (shown in Figure~\ref{fig1}). Based on this assumption, $p_k$ can be regarded as a sample from the MGD with mean $\mu_k$ and diagonal covariance $diag(\sigma_{k}^2)$, \emph{i.e.}, $N(\mu_k, diag(\sigma_{k}^2))$. Likewise, ${\hat p}_k$ is a sample from $N({\hat \mu}_k, diag(\hat{\sigma}_{k}^2))$ with mean ${\hat \mu}_k$ and diagonal covariance $diag(\hat{\sigma}_{k}^2)$.  As shown in Figure \ref{fig3_4_b}, from the view of Bayesian estimation, we regard the distribution $N({\hat \mu}_k, diag(\hat{\sigma}_{k}^2))$ as a prior, and treat the distribution $N(\mu_k, diag(\sigma_{k}^2))$ as the conditional likelihood of observed few labeled samples. Then, the Beyesian estimation of fused prototype can be expressed as their product, \emph{i.e.}, a posterior MGD $N({\hat \mu}'_k, diag({\sigma'_{k}}^2))$ with mean ${\hat \mu}'_k=\frac{\sigma_{k}^2 \odot \hat{\mu}_{k} + \hat{\sigma}_{k}^2 \odot \mu_{k}}{\hat{\sigma}_{k}^2 + \sigma_{k}^2}$ and diagonal covariance $diag({\sigma'_{k}}^2)=diag(\frac{\sigma_{k}^2 \odot \hat{\sigma}_{k}^2}{\hat{\sigma}_{k}^2 + \sigma_{k}^2})$, where $\odot$ is element-wise product (Please refer to the supplementary materials for its derivations). Finally, we take the mean $\mu'_{k}$ as the fused prototype $\hat{p}'_{k}$ to solve the few-shot tasks. We can see that ${\hat \mu}'_k$ is determined by four unknown variables $\mu_{k}$, $\sigma_{k}$, $\mu'_{k}$, and $\sigma'_{k}$. Next, we disscuss how to estimate them. 

Inspired by transductive FSL \cite{YaohuiWang_pr}, we propose to estimate the four variables by leveraging the unlabeled samples. First, we calculate the probability of each sample $x \in \mathcal{S} \cup \mathcal{Q} $ belonging to class $k$ by regarding $p_k$ and $\hat{p}_{k}$ as the prototype, respectively. For example, when we take $p_k$ as the prototype, the probability of each unlabeled sample $x \in  \cal Q $ can be computed as: 
\begin{equation}
P(y=k|x) = \frac{e^{d(f_{\theta_f}(x),\ p_k)\ \cdot\ \lambda}}
{\sum_{c} e^{d(f_{\theta_f}(x),\ {p}_c)\ \cdot\ \lambda}},
\label{eq12}
\end{equation}
where $d()$ indicates the cosine similarity of two vectors and $\lambda$ is a hyper-parameter. Following \cite{ChenLKWH19}, $\lambda=10$ is used. As for each labeled sample $x \in \mathcal{S}$, the probability turns into a one-hot vector by its labels. 
$\hat{P}(y=k|x)$ can be computed in a similar manner by using prototypes $\hat{p}_{k}$.
Second, we take $P(y=k|x)$ as sample weights and estimate the mean $\mu_{k}$ and the diagonal covariance $diag(\sigma_{k}^2)$ of each prototype distribution in a weighted average manner. That is,
\begin{small}
	\begin{equation} 
	\begin{aligned}
	\mu_{k} = \frac{1}{\sum \limits_{x \in \mathcal{S} \cup \mathcal{Q}}P(k|x)} \sum_{x \in \mathcal{S} \cup \mathcal{Q}} P(k|x) f_{\theta_f}(x),
	\end{aligned}
	\label{eq13}
	\end{equation}
\end{small}
\begin{small}
	\begin{equation} 
	\begin{aligned}
	\sigma_{k} = \sqrt{\frac{1}{\sum \limits_{x \in \mathcal{S} \cup \mathcal{Q}}P(k|x)} \sum_{x \in \mathcal{S} \cup \mathcal{Q}} P(k|x) (f_{\theta_f}(x) - \mu_{k})^2}.
	\end{aligned}
	\label{eq14}
	\end{equation}
\end{small}Then, the mean $\hat{\mu}_{k}$ and the diagonal covariance $diag(\hat{\sigma}_{k}^2)$ can be calculated in a similar manner by regarding $\hat{P}(y=k|x)$ as sample weights. In this paper, we term the overall Bayesian estimation procedure as Gaussian-based prototype fusion strategy (GaussFusion). 
\section{Performance Evaluation}

\subsection{Datasets and Settings}
\noindent \textbf{miniImagenet.} The data set is a subset of ImageNet, which includes 100 classes and each class consists of 600 images. 
Following \cite{XueW20}, we split the data set into 64 classes for training, 16 classes for validation, and 20 classes for test, respectively. The class parts/attributes are extracted from WordNet by using the relation of ``part\_holonyms()''. Note that we remove unseen parts/attributes of novel classes.

\noindent \textbf{tieredImagenet.} 
The data set is another subset of ImageNet, which includes 608 classes and each class contains about 1200 images. 
It is first partitioned into 34 high-level classes, and then split into 20 classes for training, 6 classes for validation, and 8 classes for test, respectively. Similarly, the class parts/attributes are also extracted from WordNet. 

\noindent \textbf{CUB-200-2011.}
The data set is a fine-grained classification data set, which includes 200 classes and contains about 11,788 images. 
Following \cite{zou2020compositional}, we split the data set into 100 classes for training, 50 classes for validation, and 50 classes for test, respectively. The class parts/attributes are obtained by manual annotations.

\subsection{Implementation Details}
\noindent \textbf{Architecture.}
We conduct the experiments using ResNet12 as feature extractor. In ProtoComNet, we use a single-layer perception with 256 units for the encoder, a two-layer MLP with a 300-dimensional hidden layer for the aggregator, and a two-layer MLP with 512-dimensional hidden layers for the decoder. Here, ReLU is used as activation function.

\noindent \textbf{Training Details.}
We pre-train the feature extractor with 100 epochs on base classes via an SGD with momentum of 0.9 and weight decay of 0.0005. Then, we train the ProtoComNet with 100 epochs in an episodic manner. Finally, we fine-tune all modules with 40 epochs. 

\noindent \textbf{Evaluation.} We conduct few-shot classification on 600 randomly sampled episodes from the test set and report the mean accuracy together with the 95\% confidence interval. In each episode, we randomly sample 15 query images per class for evaluation in 5-way 1-shot/5-shot tasks. 

\subsection{Discussion of Results}
For a comparison, some state-of-the-art approaches are also applied to the few-shot classification and few-shot fine-grained classification tasks as baselines. 
These methods are roughly from four types, \emph{i.e.}, metric-based, semantics-based, attribute-based, and pre-training based approaches. 

\begin{table*}[t]
	\caption{Performance on miniImagenet and tieredImagenet. The best results are shown in bold. Transductive methods are marked with *.}\smallskip
	\centering
	\smallskip\scalebox
	{0.85}{\begin{tabular}{l|c|c|c|c|c|c}
			\hline
			\multicolumn{1}{l|}{\multirow{2}{*}{Method}}&
			\multicolumn{1}{c|}{\multirow{2}{*}{Type}}& \multicolumn{1}{c|}{\multirow{2}{*}{Backbone}}&
			\multicolumn{2}{|c|}{miniImagenet} & \multicolumn{2}{|c}{tieredImagenet} \\ 
			\cline{4-7}
			& & & 5-way 1-shot & 5-way 5-shot & 5-way 1-shot & 5-way 5-shot \\
			\hline \hline
			CTM \cite{li2019finding} & Metric & ResNet18 &  62.05 $\pm$  0.55\%  & 78.63 $\pm$ 0.06$\%$ & 64.78 $\pm$ 0.11\%  & 81.05 $\pm$ 0.52\% \\
			VFSL \cite{ZhangZNXY19} & Metric & ResNet12 & $61.21 \pm 0.26\%$  & $77.69 \pm 0.17\%$ & $- \pm -\%$  & $ - \pm -\%$ \\
			RestoreNet \cite{XueW20} & Metric & ResNet18 &  $59.28 \pm 0.20\%$  & $- \pm -\%$ & $- \pm -\%$  & $- \pm -\%$ \\
			SRestoreNet$^*$ \cite{XueW20} & Metric & ResNet18 &  $61.14 \pm 0.22\%$  & $- \pm -\%$ & $- \pm -\%$  & $- \pm -\%$ \\
			\hline
			TriNet \cite{ChenFZJXS19} & Semantics & ResNet18 &  $58.12 \pm 1.37\%$  & $76.92 \pm 0.69\%$ & $- \pm -\%$  & $- \pm -\%$ \\
			AM3-PNet \cite{xing2019adaptive} & Semantics & ResNet12 &  $65.21 \pm 0.30\%$  & $75.20 \pm 0.27\%$ & $67.23 \pm 0.34\%$  & $78.95 \pm 0.22\%$ \\
			AM3-TRAML \cite{boostingfew} & Semantics & ResNet12 &  67.10 $\pm$ 0.52 $\%$  & 79.54 $\pm$ 0.60\% & $- \pm -\%$  & $- \pm -\%$ \\
			MultiSem \cite{babysteps2019} & Semantics & Dense-121 &  $67.3\%$  & \textbf{82.1}$\%$ & $- \pm -\%$  & $- \pm -\%$ \\
			FSLKT \cite{zhimao2019few} & Semantics & ConvNet128 & $64.42 \pm 0.72\%$  & $74.16 \pm 0.56\%$ & $- \pm -\%$  & $- \pm -\%$ \\
			\hline
			CPDE \cite{zou2020compositional} & Attribute & ResNet12 &  63.21 $\pm$ 0.78\%  & 79.68 $\pm$ 0.82$\%$ & $- \pm -\%$  & $- \pm -\%$ \\
			CFA \cite{Hu2019Weaklys} & Attribute & ResNet18 &  $58.50 \pm  0.80\%$  & $76.60 \pm 0.60\%$ & $- \pm -\%$  & $- \pm -\%$ \\
			\hline
			BD-CSPN$^*$ \cite{YaohuiWang_pr} & Pre-training & ResNet12 &  65.94$\%$  & 79.23$\%$ & 76.17$\%$  & 85.70$\%$ \\
			MetaBaseline \cite{chen2020new} & Pre-training & ResNet12 &  $63.17 \pm 0.23\%$  & 79.26 $\pm$ 0.17\% & $68.62 \pm 0.27\%$  & $83.29 \pm 0.18\%$ \\
			\hline
			Our Method$^*$ & Pre-training & ResNet12 & \textbf{73.13} $\pm$ \textbf{0.85}$\%$ & \textbf{82.06} $\pm$ \textbf{0.54}$\%$ & \textbf{81.04} $\pm$ \textbf{0.89}$\%$ & \textbf{87.42} $\pm$ \textbf{0.57}$\%$ \\
			\hline
	\end{tabular}}
	\label{table1}
\end{table*}

\begin{table}[t]
	\caption{Performance on CUB-200-2011. 
		The best results are shown in bold. Transductive methods are marked with *. 
	}\smallskip
	\centering
	\smallskip\scalebox
	{0.85}{\begin{tabular}{l|c|c}
			\hline
			\multicolumn{1}{l|}{\multirow{2}{*}{Method}} & \multicolumn{2}{|c}{CUB-200-2011}\\ 
			\cline{2-3}
			& 5-way 1-shot & 5-way 5-shot\\
			\hline \hline
			RestoreNet \cite{XueW20} & $74.32 \pm 0.91\%$  & $- \pm -\%$ \\
			SRestoreNet$^*$ \cite{XueW20} & $76.85 \pm 0.95\%$  & $- \pm -\%$ \\
			\hline
			TriNet \cite{ChenFZJXS19} &$69.61 \pm 0.46\%$  & $84.10 \pm 0.35\%$ \\
			MultiSem \cite{babysteps2019} & $76.1\%$  & $82.9\%$ \\ 
			\hline
			CPDE \cite{zou2020compositional} & 80.11 $\pm$ 0.34 $\%$  & 89.28 $\pm$ 0.33$\%$ \\
			CFA \cite{Hu2019Weaklys} & $73.90 \pm 0.80\%$  & $86.80 \pm 0.50\%$ \\
			\hline
			BD-CSPN$^*$ \cite{YaohuiWang_pr} & 84.90$\%$  & 90.22$\%$ \\
			\hline
			Our Method$^*$ & \textbf{93.20} $\pm$ \textbf{0.45}$\%$ & \textbf{94.90} $\pm$ \textbf{0.31}$\%$ \\
			\hline
	\end{tabular}}
    \vspace{-10pt}
	\label{table2}
\end{table}

\noindent {\bf In few-shot classification.} Table~\ref{table1} shows the results of our method and the baseline methods on miniImagenet and tieredImagenet. It can be found that our method outperforms the state-of-the-art methods, by around 2\% $\sim$ 9\%. Compared with the metric-based approaches, our method better exploits the power of pre-training by learning to complete prototypes. The results show our method is more effective, with an improvement of 4\% $\sim$ 16\%. It worth noting that our method also beats RestoreNet and SRestoreNet, which also adopt the strategy of prototype learning. This demonstrates our designed prototype completion is more effective. As for the semantics and attribute-based approaches, they also leverage the external knowledge. However, our method utilizes the knowledge to learn to complete prototypes, instead of to combine modality or to learn the feature extractor. The result validates the superiority of our manner to incorporate the external knowledge. Note that our method achieves competitive performance with the MultiSem method on 5-shot tasks on miniImagenet. We would like to emphasize that this is because MultiSem leverages a more complex backbone, namely the Dense-121 with 121 layers, instead of ResetNet12 in our model. Finally, from the results of the pre-training based apporaches, we have the following observations. (\romannumeral1) Our method outperforms BD-CSPN, by around 2\% $\sim$ 8\%. The DB-SCPN method also introduces unlabeled samples, but they only focus on pre-training and ignore the advantange of meta-learning. Different from it, we introduce a meta-learner, learning to complete prototypes, to explore the power of pre-training further. (\romannumeral2) Our method exceeds the MetaBaseline method by a large margin, around 10\%$\sim$13\% (1-shot) and 2\% $\sim$ 4\% (5-shot). This verifies our motivation that estimating more accurate prototypes is more effective than fine-tuning feature extractor during meta-learning. Besides, the improvement of performance on 1-shot tasks is more obvious than on 5-shot tasks. This is reasonable because the problem of inaccurate estimation of prototypes on 1-shot is more remarkable than 5-shot tasks.

\noindent {\bf In few-shot fine-grained classification.} Table~\ref{table2} summarizes the results on CUB-200-2011, which lead to similar observations as those in Table~\ref{table1}. We observe that our method (\romannumeral1) also achieves superior performance over state-of-the-art methods with an improvement of 5\% $\sim$ 9\%; (\romannumeral2) obtains almost consistent performance on 1-shot and 5-shot tasks, while the improvements on 1-shot task over baselines are more significant than on 5-shot. This further verifies the effectiveness of our method, especially for 1-shot tasks.

\subsection{Statistical Analysis}
\begin{figure}[t]
	\centering
	\includegraphics[width=0.90\columnwidth]{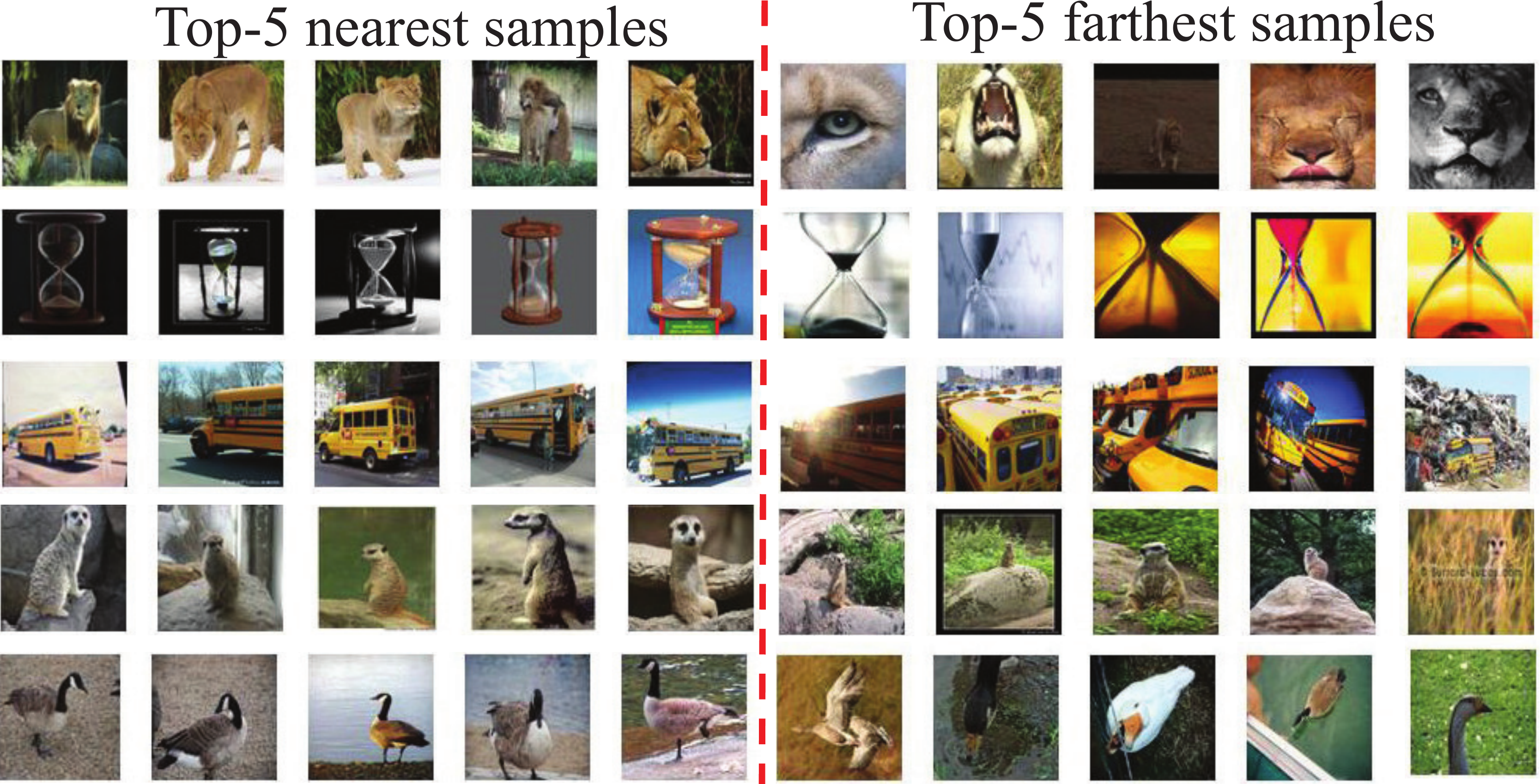}  
	\caption{Top-5 nearest and farthest samples from centers.}
	\label{fig4}
\end{figure}

\noindent {\bf Is our idea reasonable on realistic data?} We randomly select five classes from the novel classes of miniImageNet and retrieve top-5 nearest and farthest samples from its ground-truth class center in the feature space. As shown in Figure~\ref{fig4}, the nearest images are more complete; however, the farthest samples are missing partial parts/attributes due to its incompleteness, noise background, or obscured details.

\begin{table}
	\caption{The cosine similarity between the estimated and real prototypes.
		$d(x, y)$ denotes the cosine simiarity of vectors $x$ and $y$.}\smallskip
	\centering
	\smallskip\scalebox
	{0.85}{\begin{tabular}{l|c|c|c}
			\hline
			Methods & $d(p_k, p^{real}_k)$ & $d(\hat{p}_k, p^{real}_k)$ & $d(\hat{p}'_k, p^{real}_k)$\\
			\hline \hline
			SRestoreNet  & 0.55 & 0.78 & 0.79\\
			FSLKT  & 0.55 & - & 0.68\\
			BD-CSPN  & 0.55 & - & 0.67\\
			Our Method & 0.55 & 0.71 & 0.90\\
			\hline
	\end{tabular}}
	\vspace{-10pt}
	\label{table3}
\end{table} 

\noindent {\bf Does our method obtain more accurate prototypes?} We calculate the average cosine similarity between the estimated prototypes and the real prototypes on 1000 episodes (5-way 1-shot) of miniImagenet. Three results including the mean-based ($p_k$), the restored/completed ($\hat{p}_k$) and the fused prototype ($\hat{p}'_k$) are reported. For a fair comparison, we report the results of SRestoreNet, FSLKT, and BD-CSPN as the baselines. As shown in Table~\ref{table3}, the results show that our method obtains more accurate prototypes than these baselines. Note that the prototype $\hat{p}_k$ from SRestoreNet is better than our method. This is reasonable because they leverage unlabeled samples before restoring prototypes. However, we exploit them after completing prototypes. 

\begin{figure}[t]
	\centering
	\includegraphics[width=0.90\columnwidth]{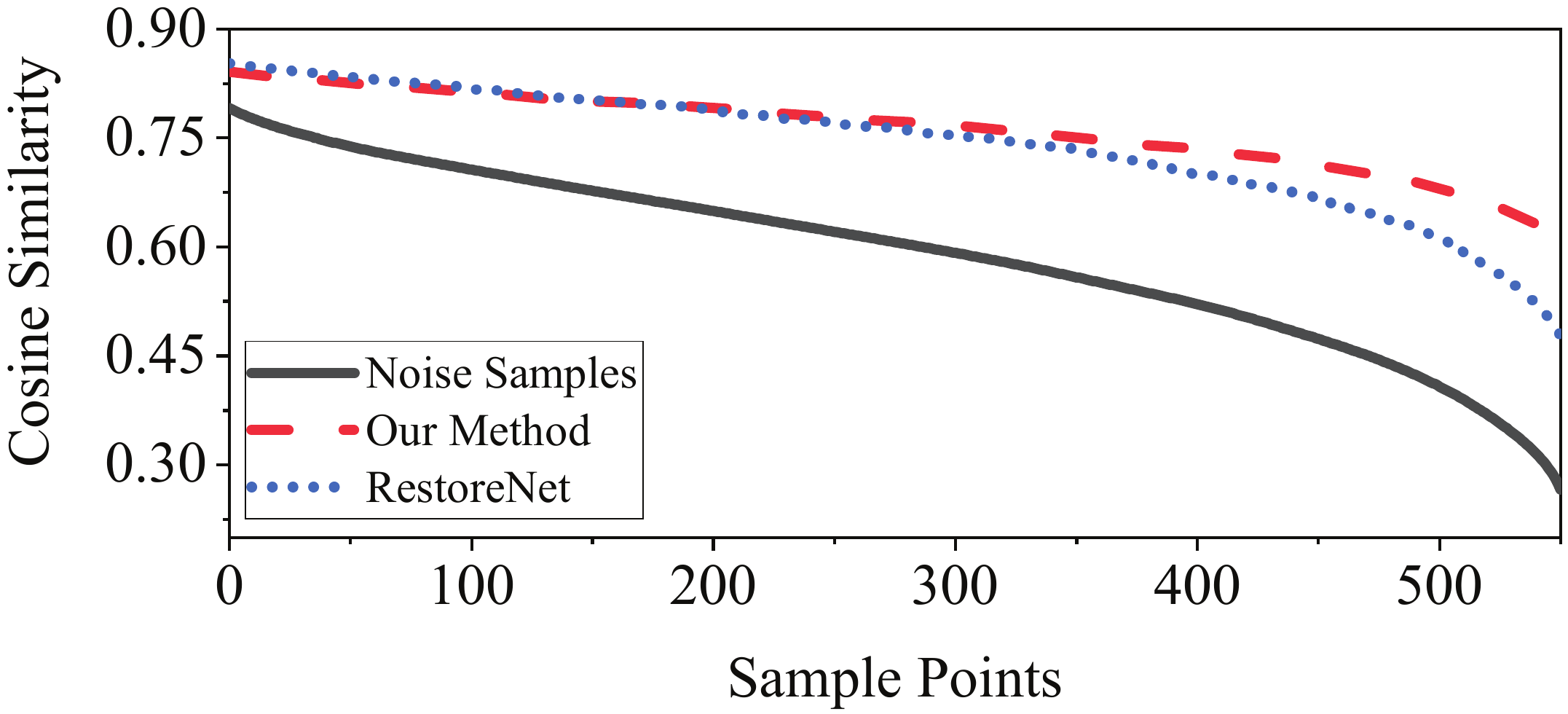} 
	\caption{Performance analysis of ProtoComNet.}
	\label{fig2_0}
\end{figure}

\noindent {\bf Is our method effective for the samples far away from its class center?} On the novel classes of miniImageNet, we calculate the cosine similarity between each noise image and its class center and sort them in descending order (\emph{i.e.}, the larger the sample number is, the farther away it is from the class center). Then, we take the noise images as inputs to predict the prototypes by using our method and RestoreNet, respectively. The cosine similarity between predicted prototypes and real class centers is shown in Figure~\ref{fig2_0}. Note that (\romannumeral1) we smoothen the curve through moving average with 50 samples; (\romannumeral2) we show the average results for all novel classes. We observe our method achieves more accurate prototypes than RestoreNet and the improvement becomes larger as the samples are farther away from its center. 

\subsection{Ablation Study}
We conduct an ablation study on miniImagenet, to assess the effects of the two specially-designed components, \emph{i.e.}, learning to complete prototypes and Gaussian-based prototype fusion strategy. Specifically, (\romannumeral1) we remove all components, \emph{i.e.}, classifying each sample by the mean-based prototypes; (\romannumeral2) we add the ProtoComNet on (\romannumeral1) and classify each sample by the completed prototypes; (\romannumeral3) we average the mean-based and completed prototypes to obtain the final prototypes, which is the fusion strategy in \cite{XueW20}; (\romannumeral4) we replace the prototype fusion strategy of (\romannumeral3) by our GaussFusion. The results are shown in Table~\ref{table4}. 

\noindent {\bf Learning to Complete Prototypes.} From the results of the first and second row in Table~\ref{table4}, we observe that 1) the latter exceeds the former in 1-shot tasks, by around 4\%, which means that learning to complete prototypes is effective; 2) the latter obtains poor performance in 5-shot tasks. As our analysis in Section~\ref{section3_4}, the phenomenon results from the bias of ProtoComNet, namely the primitive knowledge noises or base-novel class differences. 

\noindent {\bf Gaussian-based Prototype Fusion Strategy.} According to the result in the last three rows of Table~\ref{table4}, we find that 1) the problem of ProtoComNet with poor performance on 5-shot tasks is effectively solved after we add mean-based prototype fusion strategy; 2) the performance of the ProtoComNet can be further improved when it is combined with GaussFusion, by around 3\% on classification accuracy. The result suggests that the GaussFusion is more effective than the mean-based fusion strategy. The key reason is GaussFussion effectively estimates prototype distribution by exploiting the unlabelled samples. To further vertify that GaussFusion is able to alleviate the prototype completion error problem, we analyze the impacts of primitive knowledge with different noise levels $\gamma$ on classification performance in Table~\ref{table5}. Here, we introduce noises by randomly adding or removing class parts/attributes with probability $\gamma$. It can be observed that our method is more robust to primitive knowledge noises when GaussFusion is applied.

\begin{table}[t]
		\caption{Ablation study on miniImagenet. LCP: Learning to complete prototypes. GF, MF: Gaussian, mean-based prototype fusion.}\smallskip
	\centering
	\smallskip\scalebox
	{0.85}{\begin{tabular}{c|c|c|c|c|c}
			\hline
			 & LCP & GF & MF & 5-way 1-shot & 5-way 5-shot \\
			\hline \hline
			(\romannumeral1) & & &&61.22 $\pm$ 0.84$\%$ & 78.72 $\pm$ 0.60$\%$ \\
			(\romannumeral2) &$\surd$ & && 65.62 $\pm$ 0.79$\%$ & 75.32 $\pm$ 0.61$\%$ \\
			(\romannumeral3) & $\surd$ &  & $\surd$ & 70.14 $\pm$ 0.81$\%$ & 79.70 $\pm$ 0.60$\%$\\
			(\romannumeral4) & $\surd$ & $\surd$ && 73.13 $\pm$ 0.85$\%$ & 82.06 $\pm$ 0.54$\%$ \\
			\hline
	\end{tabular}}
	\label{table4}
\end{table}

\begin{table}
	\caption{The performance analysis of primitive knowledge with different noise level $\gamma$ on 5-way 1-shot tasks of miniImagenet.}\smallskip
	\centering
	\smallskip\scalebox
	{0.80}{\begin{tabular}{l|c|c|c|c}
			\hline
			Methods & $\gamma=0.0$ & $\gamma=0.1$ & $\gamma=0.2$ & $\gamma=0.3$\\
			\hline \hline
			w/o GaussFusion  & 65.62 $\%$ & 59.28 $\%$ & 55.39 $\%$ & 51.97 $\%$ \\
			w/ GaussFusion  & 73.13 $\%$ & 71.80 $\%$ & 70.77 $\%$ & 69.97 $\%$\\
			\hline
	\end{tabular}}
	\vspace{-10pt}
	\label{table5}
\end{table} 

\section{Conclusions}
For few-shot learning, a simple pre-training on base classes can obtain a good feature extractor, where the novel class samples can be well clustered together. The key challenge is how to obtain more representative prototypes because the novel class samples spread as groups with large variances. To solve the issue, we propose a prototype completion network to complete prototypes via primitive knowledge, and a Gaussian-based prototype fusion strategy to alleviate the prototype completion error problem. Experiments show that our method obtains superior performance on three data sets. In the future, we are interested in exploring more efficient attribute modeling strategy such as incorporating unseen parts/attributes into our framework, so that more accurate prototypes can be delivered for novel classes.

\section*{Acknowledgments} This work was supported by the Shenzhen Science and Technology Program under Grant No. JCYJ201805071- 83823045 and Grant No. JCYJ20200109113014456.

{\small
\bibliographystyle{ieee_fullname}
\bibliography{egbib}
}

\section*{Supplementary Material}
\noindent {\bf Proposition.} Let $f(x)$ and $g(x)$ be a Multivariate Gaussian Distributions with diagonal covariance, \emph{i.e.}, $f(x) = N(\hat{\mu}_{k}, diag(\hat{\sigma}_{k}^2))$ and $g(x) = N(\mu_{k}, diag(\sigma_{k}^2))$ where $x$ is a $d$-dimension random vector, $\hat{\mu}_{k}$ and $\mu_{k}$ denote $d$-dimension mean vector, and $\hat{\sigma}_{k}^2$ and $\sigma_{k}^2$ are $d$-dimension variance vector. Then, their product obeys a new Multivariate Gaussian Distributions $N(\mu'_{k}, diag({\sigma'_{k}}^2))$ with $\mu'_{k}=\frac{\sigma_{k}^2 \odot \hat{\mu}_{k} + \hat{\sigma}_{k}^2 \odot \mu_{k}}{\hat{\sigma}_{k}^2 + \sigma_{k}^2}$ and ${\sigma'_{k}}^2=\frac{\sigma_{k}^2 \odot \hat{\sigma}_{k}^2}{\hat{\sigma}_{k}^2 + \sigma_{k}^2}$, where $\odot$ denotes the element-wise product.

\noindent {\bf Derivation.} Considering that the covariances of $f(x)$ and $g(x)$ are simplified as diagonal covariances. This means that the variables of the random vector $x$ are uncorrelated. In this case, $f(x)$ and $g(x)$ can be simplified as the expression below:

\begin{equation} 
\nonumber 
f(x)=\prod_{i=0}^{d-1} \frac{1}{\sqrt{2\pi{\hat{\sigma}_{k,i}}^2}}\ e^{(\frac{-(x_i-\hat{\mu}_{k,i})^2}{2\hat{\sigma}_{k,i}^2})}
\end{equation}
\begin{equation} 
\nonumber 
g(x)=\prod_{i=0}^{d-1} \frac{1}{\sqrt{2\pi{\sigma^2_{k,i}}}}\ e^{(\frac{-(x_i-\mu_{k,i})^2}{2\sigma_{k,i}^2})}
\end{equation}
Thus, their product $h(x)$ satisfies:
\begin{equation}
\nonumber 
\begin{aligned}
&h(x) \\&= f(x)g(x) 
\\& = \prod_{i=0}^{d-1} \frac{1}{\sqrt{2\pi\hat{\sigma}^2_{k,i}}}\ e^{(\frac{-(x_i-\hat{\mu}_{k,i})^2}{2\hat{\sigma}_{k,i}^2})} \ \frac{1}{\sqrt{2\pi\sigma^2_{k,i}}}\ e^{(\frac{-(x_i-\mu_{k,i})^2}{2\sigma_{k,i}^2})}
\\& = \prod_{i=0}^{d-1} \frac{1}{2\pi\sqrt{\hat{\sigma}^2_{k,i}\sigma^2_{k,i}}}\ e^{(\frac{-(x_i-\hat{\mu}_{k,i})^2}{2\hat{\sigma}_{k,i}^2} + \frac{-(x_i-\mu_{k,i})^2}{2\sigma_{k,i}^2})}
\\& = \prod_{i=0}^{d-1} \frac{1}{2\pi\sqrt{\hat{\sigma}^2_{k,i}\sigma^2_{k,i}}}\ e^{(\frac{(x_i-\frac{\sigma_{k,i}^2\hat{\mu}_{k,i} + \hat{\sigma}_{k,i}^2\mu_{k,i}}{\hat{\sigma}_{k,i}^2 + \sigma_{k,i}^2})^2}{2\frac{\sigma_{k,i}^2 \hat{\sigma}_{k,i}^2}{\hat{\sigma}_{k,i}^2 + \sigma_{k,i}^2}} + \frac{(\hat{\mu}_{k,i} - \mu_{k,i})^2}{2(\hat{\sigma}_{k,i}^2 + \sigma_{k,i}^2)})}
\\&= \prod_{i=0}^{d-1} \frac{S_{i}}{\sqrt{2\pi\frac{\sigma_{k,i}^2 \hat{\sigma}_{k,i}^2}{\hat{\sigma}_{k,i}^2 + \sigma_{k,i}^2}}} e^{(-\frac{-(x_i-\frac{\sigma_{k,i}^2\hat{\mu}_{k,i} + \hat{\sigma}_{k,i}^2\mu_{k,i}}{\hat{\sigma}_{k,i}^2 + \sigma_{k,i}^2})^2}{2(\frac{\sigma_{k,i}^2 \hat{\sigma}_{k,i}^2}{\hat{\sigma}_{k,i}^2 + \sigma_{k,i}^2})})}
\end{aligned}
\end{equation}
where $S_{i}=\frac{1}{\sqrt{2\pi(\sigma_{k,i}^2 + \hat{\sigma}_{k,i}^2)}} e^{-\frac{(\hat{\mu}_{k} - \mu_{k})^2}{2(\hat{\sigma}_{k}^2 + \sigma_{k}^2)}}$. Thus, $h(x)$ is also a multivariate Gaussian distribution, \emph{i.e.}, $N(\mu'_{k}, diag({\sigma'_{k}}^2))$ with mean $\mu'_{k}=\frac{\sigma_{k}^2 \odot \hat{\mu}_{k} + \hat{\sigma}_{k}^2 \odot \mu_{k}}{\hat{\sigma}_{k}^2 + \sigma_{k}^2}$ and diagonal covariance $diag({\sigma'_{k}}^2)$ where ${\sigma'_{k}}^2=\frac{\sigma_{k}^2 \odot \hat{\sigma}_{k}^2}{\hat{\sigma}_{k}^2 + \sigma_{k}^2}$.

\end{document}